# InstantID: Zero-shot Identity-Preserving Generation in Seconds

Qixun Wang[1,2], Xu Bai[1,2], Haofan Wang[1,2]⋆, Zekui Qin[1,2], Anthony Chen[1,2,3], Huaxia Li[2], Xu Tang[2], and Yao Hu[2]

InstantX Team[1], Xiaohongshu Inc[2], Peking University[3]
{haofanwang.ai@gmail.com}
https://instantid.github.io

**Abstract.** There has been significant progress in personalized image synthesis with methods such as Textual Inversion, DreamBooth, and LoRA. Yet, their real-world applicability is hindered by high storage demands, lengthy fine-tuning processes, and the need for multiple reference images. Conversely, existing ID embedding-based methods, while requiring only a single forward inference, face challenges: they either necessitate extensive fine-tuning across numerous model parameters, lack compatibility with community pre-trained models, or fail to maintain high face fidelity. Addressing these limitations, we introduce **InstantID**, a powerful diffusion model-based solution. Our plug-and-play module adeptly handles image personalization in various styles using just a single facial image, while ensuring high fidelity. To achieve this, we design a novel IdentityNet by imposing strong semantic and weak spatial conditions, integrating facial and landmark images with textual prompts to steer the image generation. InstantID demonstrates exceptional performance and efficiency, proving highly beneficial in real-world applications where identity preservation is paramount. Moreover, our work seamlessly integrates with popular pre-trained text-to-image diffusion models like SD1.5 and SDXL, serving as an adaptable plugin. Our codes and pre-trained checkpoints will be available at https://github.com/InstantID/InstantID.

## 1 Introduction

Image generation technology has seen remarkable advancements with the emergence of large text-to-image diffusion models such as GLIDE[12], DALL·E 2[15], Imagen[19], and Stable Diffusion (SD)[16]. A focal point of these developments is personalized and customized generation, which seeks to create images consistent in style, subject, or character ID based on one or more reference images. This approach has garnered significant interest for its broad potential in applications like E-commerce advertising, AI portraits, image animation, and virtual try-ons. However, a notable challenge remains: generating customized images that accurately preserve the intricate identity details of human subjects. This task is particularly demanding, as human facial identity (ID) involves more nuanced semantics and requires a higher standard of detail and fidelity compared to general styles or objects, which primarily focus on coarse-grained textures and colors. Existing vanilla text-to-image models, dependent on detailed textual descriptions, fall short in achieving strong semantic relevance in customized generation. To enhance controllability, recent works such as ControlNet[25] and T2I-adapter[11] have been introduced. These advancements add spatial conditioning controls to large pre-trained text-to-image models, facilitating fine-grained structural control through elements like user-drawn sketches, depth maps, body poses, and semantic segmentation maps. Additionally, adaptations for style or content referencing, such as the T2I-adapter's style adapter and Uni-ControlNet's[26] global controller, have made strides in this area. Despite these advancements, the fidelity of the generated images to the reference remains only partial.

---

⋆ Corresponding author

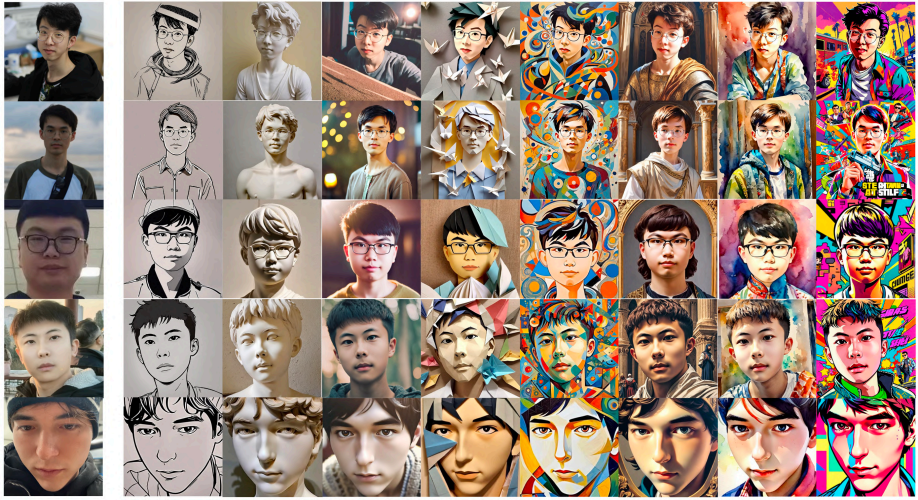

**Fig. 1.** The awesome team of InstantID in author order, generated with InstantID.

To enhance image generation with reference images, current personalized generation methods can be classified into two types, based on the need for fine-tuning during testing. The first type includes methods necessitating fine-tuning, with leading examples like DreamBooth[18], Textual Inversion[4], and Low-Rank Adaptation[6] (LoRA). These methods aim to fine-tune pre-trained text-to-image models to better reflect new aspects in reference images. While effective in achieving high accuracy, the fine-tuning process is resource-intensive and time-consuming, limiting their practicality. Additionally, they often require multiple references in various settings and struggle in limited-data scenarios, such as with identical character IDs.

The second type bypasses fine-tuning during inference. It involves constructing a large amount of domain-specific data and building a lightweight adapter for feature extraction from the reference image. This adapter then integrates these features into the diffusion generation process using cross-attention. IP-Adapter[24] is one of the representative works, which employs a unique cross-attention mechanism to separate text and image features, allowing for the injection of the reference image as a visual prompt. However, this approach, which primarily relies on CLIP[14]'s image encoder, tends to produce only weakly aligned signals, falling short in creating high-fidelity, customized images.

In light of these constraints, we introduce a novel approach (**InstantID**) focusing on instant identity-preserving image synthesis. In this study, we bridge the gap between high fidelity and efficiency by introducing a simple plug-and-play module, enabling it to adeptly handle image personalization in any style using only one facial image while maintaining high fidelity. To preserve the facial identity from the reference image, we design a novel face encoder to retain the intricate details by adding strong semantic and weak spatial conditions, which incorporate facial image, landmark image and textual prompts to guide the image generation process. We distinguish our work from previous ones in the following aspects: (1) Plugability and compatibility: we focus on training a lightweight adapter instead of the full parameters of UNet[17], making our module pluggable and compatible with pre-trained models in the community; (2) Tuning-free: our method requires only a single forward propagation for inference, eliminating the need for fine-tuning. This feature makes InstantID highly economical and practical for real-world applications; (3) Superior performance: with just one reference image, InstantID achieves state-of-the-art results, demonstrating high fidelity and flexibility. Remarkably, it can match or even surpass the performance of training-based methods like LoRAs, which rely on multiple reference images.

In summary, our contributions are threefold:

• We present InstantID, an innovative ID-preserving adaptation method for pre-trained text-to-image diffusion models to decently bridge the gap between fidelity and efficiency. Experimental results demonstrate the excellent performance of our proposed method as compared with other state-of-the-art methods in this domain.

• InstantID is pluggable and compatible with other custom models fine-tuned from the same base diffusion model, enabling ID preservation in pre-trained models at no additional cost. Furthermore, InstantID maintains considerable control over text editing as observed in the original Stable Diffusion model, enabling the smooth integration of IDs into various styles.

• The excellent performance and efficiency of InstantID ignite its huge potential for a range of real-world applications, such as novel view synthesis, ID interpolation, multi-ID and multi-style synthesis.

## 2  Related Work

### 2.1  Text-to-image Diffusion Models

Text-to-image diffusion models [15,19,16,12,1,7] achieve state-of-the-art image generation results and have received unprecedented interest from the community in recent years. A common practice is to encode the text prompt into latent through a pre-trained language encoder like CLIP [14] and use the latent to guide the diffusion process. For instance, GLIDE [12] adopts a cascaded text-guided diffusion architecture to support both image generation and editing. DisCo [21] uses the text encoder from CLIP [14] to process text inputs and introduces CLIP guidance in the diffusion process. Stable Diffusion [16] is one of the highlighted works of latent diffusion models (LDM) where the diffusion process is performed in the latent image space instead of the original pixel space, which significantly reduces the computation cost. As a follow-up, Stable Diffusion XL [13] (SDXL) is a powerful text-to-image generation model that introduced a larger UNet and a second text encoder to gain stronger textual control over generated images.

### 2.2  Subject-driven Image Generation

Subject-driven text-to-image generation, which uses a limited set of images of a particular subject to generate customized images based on text descriptions, has seen notable advancements. Previous subject-driven approaches like DreamBooth [18], Textual Inversion [4], ELITE [22], E4T [5], and ProFusion [28] fine-tune a special prompt token $S^*$ to describe the target concepts during the fine-tuning process. In contrast, other recent methods strive to enable subject-driven text-to-image generation without the need for additional fine-tuning. These methods typically involve training additional modules while keeping the core pre-trained text-to-image models frozen. A leading example of this approach is IP-Adapter [24], which aims to decouple the cross-attention mechanism by separating the cross-attention layers for text features and image features. A concurrent work, Anydoor [2], complements the commonly used identity feature with detail features that are designed to maintain texture details while allowing versatile local variations.

### 2.3  ID Preserving Image Generation

ID-preserving image generation is a special case of subject-driven generation, but it focuses on face attributes with strong semantics and finds broad application in real-world scenarios. Existing works can be mainly divided into two categories based on their reliance on test-time fine-tuning. Low-Rank Adaptation [6] (LoRA) is a popular lightweight training technique that inserts a minimal number of new weights into the model before training on customized datasets. However, LoRA necessitates individual training for each new character, limiting its flexibility. In contrast, recent developments have introduced optimization-free methods, bypassing additional fine-tuning or inversion processes. Face0 [20] overrides the last three text tokens with the projected face embedding in CLIP space and uses the joint embedding as a condition to guide the diffusion process. PhotoMaker [10] adopts a similar approach, but enhances its ability to extract ID-centric embeddings by fine-tuning part of the Transformer [3] layers in the image encoder and merging the class and image embeddings. FaceStudio [23] presents a hybrid-guidance identity-preserving image synthesis framework, where the face embedding is integrated into both CLIP vision embeddings and CLIP text embeddings via linear projection, then the merged guidance embedding is fused into the UNet with cross-attentions. IP-Adapter-FaceID [24] uses face ID embedding from a face recognition model instead of CLIP image embedding to retain ID consistency. Nevertheless, these methods either necessitate training the full parameters of UNet, sacrificing compatibility with existing pre-trained community models, or fall short in ensuring high face fidelity. To address these limitations, we introduce a pluggable module that adeptly bridges the divide between training-free and training-intensive methods. Our method, requiring no fine-tuning during inference, seamlessly aligns with readily available pre-trained diffusion models like SD1.5 and SDXL, achieving exceptional fidelity in face preservation.

## 3  Methods

InstantID is an efficient lightweight adapter, effortlessly endowing pre-trained text-to-image diffusion models with the capability of ID preservation, akin to a 'free lunch'. In this part, we first discuss preliminaries about the text-to-image diffusion model and custom image generation in Section 3.1, and then introduce the detailed design of our InstantID in Section 3.2. Finally, the specifics of our training and inference process are detailed in Section 3.3.

### 3.1  Preliminaries

**Stable Diffusion**. Our method is built on top of Stable Diffusion [16], which performs the diffusion process computationally efficiently in low-dimensional latent space instead of pixel space with an auto-encoder [8]. Specifically, given an input image $x_i \in R^{H \times W \times 3}$, the encoder first maps it to a latent representation: $z_0 = \xi(x_i)$, where $z_0 \in R^{h \times w \times c}$, $f = H/h = W/w$ is the downsampling factor and $c$ is the latent dimension. The diffusion process adopts a denoising UNet [17] $\epsilon_\theta$ to denoise a normally-distributed noise $\epsilon$ with noisy latent $z_t$, current timestep $t$ and condition $C$, where $C$ represents the embedding of textual prompts generated via a pre-trained CLIP [14] text encoder. The overall training objective is defined as

$$\mathcal{L} = \mathbb{E}_{z_t,t,C,\epsilon \sim \mathcal{N}(0,1)}[||\epsilon - \epsilon_\theta(z_t, t, C)||_2^2]. \tag{1}$$

**Controllable Diffusion Models**. We utilize ControlNet[25] as an exemplar, which is capable of adding spatial control to a pre-trained diffusion model as conditions, extending beyond the capabilities of basic textual prompts. ControlNet integrates the UNet architecture from Stable Diffusion with a trainable replica of UNet. This replica features zero convolution layers within the encoder blocks and the middle block. The full process of ControlNet is executed as follows,

$$y_c = \mathcal{F}(x, \theta) + \mathcal{Z}(\mathcal{F}(x + \mathcal{Z}(c, \theta_{z1}), \theta_c), \theta_{z2}). \tag{2}$$

ControlNet distinguishes itself from the original Stable Diffusion in the latter residual item, where $\mathcal{F}$ is the UNet architecture, $x$ is the latent, $\theta$ is the frozen weight of the pre-trained model, $\mathcal{Z}$ corresponds to zero convolutions with weights $\theta_{z1}$, $\theta_{z2}$, and $\theta_c$ is the trainable weight of ControlNet. To put it simply, ControlNet encodes spatial condition information (sketches, poses, depth maps, etc.) by adding residuals to UNet Block and then embeds it into the original network.

**Image Prompt Adapter**. IP-Adapter[24] introduces a novel approach to achieving image prompt capabilities in parallel with text prompts without modifying the original text-to-image models. It employs a unique decoupled cross-attention strategy, embedding image features through additional cross-attention layers while leaving other parameters unchanged. The decoupled cross-attention can be illustrated as

$$Z_{new} = Attention(Q, K^t, V^t) + \lambda \cdot Attention(Q, K^i, V^i), \tag{3}$$

where $Q$, $K^t$, $V^t$ are the query, key, and values matrices of the attention operation for text cross-attention, $K^i$ and $V^i$ are for image cross-attention. Given the query features $Z$ and the image features $c_i$, $Q = ZW_q$ $K^i = c_iW_k^i$, $V^i = c_iW_v^i$. Note that only $W_k^i$ and $W_k^i$ are trainable weights.

### 3.2 Methodology

**Overview**. Given only one reference ID image, InstantID aims to generate customized images with various poses or styles from a single reference ID image while ensuring high fidelity. Figure 2 provides an overview of our method. It incorporates three crucial components: (1) An ID embedding that captures robust semantic face information; (2) A lightweight adapted module with decoupled cross-attention, facilitating the use of an image as a visual prompt; (3) An IdentityNet that encodes the detailed features from the reference facial image with additional spatial control.

**ID Embedding**. Contrary to prior approaches like IP-Adapter [24], FaceStudio [23], and PhotoMaker [10], which rely on a pre-trained CLIP image encoder for visual prompt extraction, our work targets the need for stronger semantic details and enhanced fidelity in the ID preservation task. CLIP's inherent limitation lies in its training on weakly aligned data, which means its encoded features predominantly capture broad, ambiguous semantic information like composition, style, and colors. Such features can act as general supplements to text embeddings, but they fall short for tasks requiring precise ID preservation, where stronger semantics and higher fidelity are paramount. Over recent years, extensive research in face representation, particularly in face recognition and ReID, has demonstrated the efficacy of face representation in complex tasks like face recognition and reconstruction. However, its direct application in image generation with diffusion models is underexplored. In our research, we leverage a pre-trained face model to detect and extract face ID embedding from the reference facial image, providing us with strong identity features to guide the image generation process. This leads us to a critical question: **how do we effectively inject the identity features into the diffusion models?** We explore and answer this question in the subsequent sections of our work.

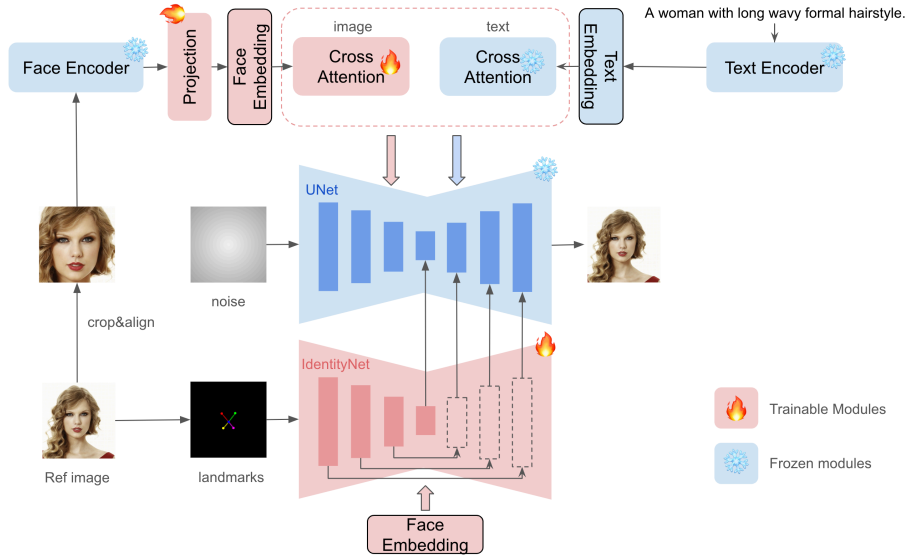

**Fig. 2. The overall pipeline of our proposed InstantID.** Our model consists of three parts to maintain high facial fidelity. First, we adopt a face encoder instead of CLIP to extract semantic face features and use a trainable projection layer to project them to the space of text features. We take the projected feature as face embedding. Then, a lightweight adaptive module with decoupled cross-attention is introduced to support images as prompts. Finally, we propose IdentityNet to encode complex features in reference facial images with additional weak spatial control. In IdentityNet, the generation process is fully guided by face embedding without any textual information. Only newly added modules are updated, while the pre-trained text-to-image model remains frozen to ensure flexibility. After training, users can generate ID-preserving images of any style in high fidelity for free.

**Image Adapter**. The capability of image prompting in pre-trained text-to-image diffusion models significantly enhances textual prompts, particularly for content that is challenging to describe with text. In our approach, we adopt a strategy similar to IP-Adapter for image prompting, as shown in Figure 2 upper branch, a lightweight adaptive module with decoupled cross-attention is introduced to support images as prompts. However, we diverge by employing ID embedding as our image prompt, as opposed to the coarse-aligned CLIP embedding. This choice is aimed at achieving a more nuanced and semantically rich prompt integration.

**IdentityNet**. Despite the availability of methods to integrate image prompts with text prompts (e.g. IP-Adapter), we argue that these techniques offer only coarse-grained improvements, and such a level of integration is insufficient for ID-preserving image generation. We attribute this limitation to the inherent training mechanism and properties of the pre-trained diffusion model. For instance, when image and text prompts are concatenated before the attention layers, the model struggles with fine-grained control over extended token sequences. Yet, directly adding the text and image tokens in cross-attention tends to weaken the control exerted by text tokens. Moreover, attempts to enhance the strength of image tokens for improved fidelity can inadvertently impair the editing abilities of text tokens. This presents a significant challenge in our work, which we address by adopting an alternative feature embedding method, ControlNet. This method typically utilizes spatial information as input for the controllable module, maintaining consistency with UNet settings in the diffusion model and including text as a conditional element in cross-attention layers.

In our adaptation of ControlNet, there are mainly two modifications: 1) Instead of fine-grained OpenPose facial keypoints, we use only five facial keypoints (two for the eyes, one for the nose, and two for the mouth) for conditional input. 2) We eliminate the text prompts and use ID embedding as conditions for cross-attention layers in the ControlNet. We discuss our motivation below.

First, we recognize that while spatial control is essential in ControlNet, it must be sensitively applied, especially for facial regions. We opt for facial landmarks as spatial control signals but restrict ourselves to five key points (two for the eyes, one for the nose, and two for the mouth) for a more generalized constraint than detailed key points. This approach serves two purposes: a) InstantID is trained on real-world human images, not manually cropped face images, where faces are often a minor part of the scene, making precise key point detection challenging. b) We aim to reduce the impact of spatial constraints and prevent overemphasis on redundant facial information, like face shape or mouth closure, to maintain editability. Nonetheless, we find that without any spatial constraints, the human face's degree of freedom may be excessively large, complicating the generation of satisfactory results. A map of coarse key points, therefore, offers a balanced solution.

Second, the reason why we eliminate the text prompts and use ID embedding as condition for cross-attention layers instead is because this approach enables the network to focus exclusively on ID-related representations, thereby not being influenced by generalized descriptions of faces and backgrounds.

### 3.3 Training and Inference Strategies

During training, we only optimize the parameters of the Image Adapter and the IdentityNet while keeping the parameters of the pre-trained diffusion model frozen. We train the entire InstantID pipeline on image-text pairs featuring human subjects, employing a training objective akin to that used in the original stable diffusion work:

$$\mathcal{L} = \mathbb{E}_{z_t,t,C,C_i,\epsilon \sim \mathcal{N}(0,1)}[||\epsilon - \epsilon_\theta(z_t, t, C, C_i)||_2^2], \tag{4}$$

where $C_i$ is the task-specific image condition on IdentityNet. Note that in the training process, we do not randomly drop text or image conditions, as we have removed the text prompts condition in IdentityNet.

A key feature of our work is the separation between text cross-attention and image cross-attention within the image prompt adapter, as well as the decoupling of the image prompt adapter and IdentityNet. This design choice allows us to independently and flexibly adjust the weight of these image conditions, ensuring a more controlled and targeted training and inference process.

## 4 Experiments

We implement with Stable Diffusion and train it on the large-scale open-source dataset LAION-Face [27], which consists of 50 million image-text pairs to ensure diversity. In addition, we collect 10 million high-quality human images from the Internet with annotations automatically generated by BLIP2 [9] to further improve the generation quality. We focus on single-person images and utilize a pre-trained face model, antelopev2[1], for detecting and extracting face ID embeddings from human images. We train on the original human images instead of cropped face datasets. While training, only the parameters of the Image Adapter and IdentityNet are updated, while the pre-trained text-to-image model remains frozen. Our experiments are based on the SDXL-1.0 model and are conducted on 48 NVIDIA H800 GPUs (80GB) with a batch size of 2 per GPU.

### 4.1 Qualitative Results

We present qualitative results in various settings to demonstrate the robustness, prompt editability, and compatibility of our method.
**Image Only**. We use an empty prompt to guide the generation process entirely by the reference image. Our results without prompts, as seen in Figure 3 'empty prompt', demonstrate InstantID's ability to robustly maintain rich semantic facial content like expression, age, and identity. However, certain semantics, such as gender, may not be perfectly accurate (see second row).

---

[1] https://github.com/deepinsight/insightface

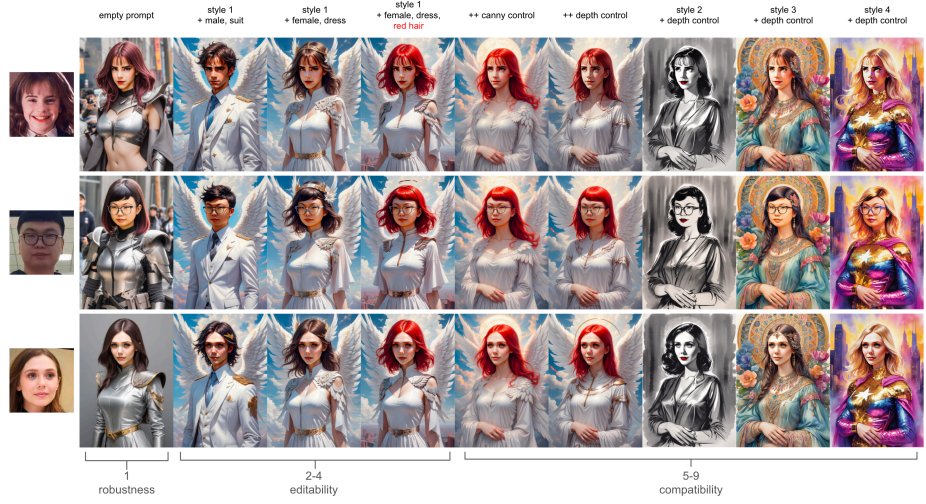

**Fig. 3. Demonstration of the robustness, editability, and compatibility of InstantID.** Column 1 shows the result of Image Only results where the prompt is set to empty during inference. Columns 2-4 show the editability through text prompt. Columns 5-9 show the compatibility with existing ControlNets (canny & depth).

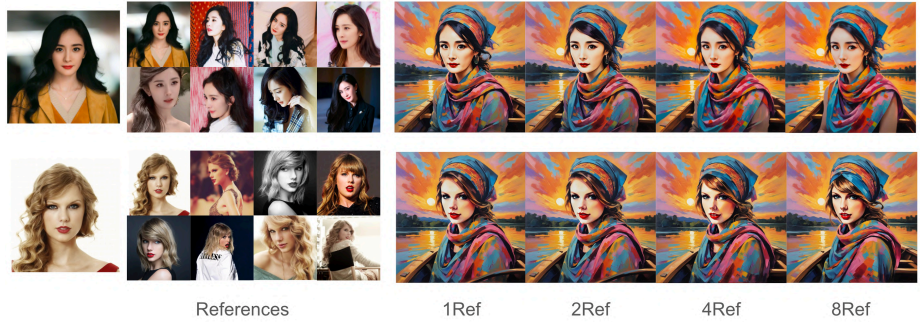

**Fig. 4. Effect of the number of reference images.** For multiple reference images, we take the average mean of ID embeddings as image prompt. InstantID is able to achieve good results even with only one single reference image.

**Image + Prompt**. In several prompt settings (Figure 3, columns 2-4), we observe no degradation in text control capabilities while ensuring identity consistency, effectively changing gender, clothes, and hair color.

**Image + Prompt + Spatial Control**. Our method is compatible with pre-trained spatial control models like ControlNet. Figure 3, columns 5-9, illustrates this compatibility, showing that we can introduce flexible spatial control using pre-trained ControlNet (canny, depth) models.

### 4.2  Ablative Study

We assess the effectiveness of each internal module during inference and its impact on the generated results. Figure 9 in Appendix demonstrates that IdentityNet alone achieves good ID retention, and the addition of the Image Adapter further enhances facial detail restoration.

We also examine the influence of the number of reference images on generation quality in Figure 4. More reference images generally improve our generation quality, but even with a single image, our method achieves remarkable fidelity. Note that in training-based methods, the number of images of the same person usually directly affects the quality of generation. For example, LoRAs training usually requires dozens of images, and if the image similarity is too high, it will be troubled by over-fitting. Yet in our case, we use the average embedding of all reference images, and this further helps improve our generation quality.

### 4.3  Comparison to Previous Methods

**With IP-Adapters.** To demonstrate the effectiveness of our method, we compare our work with existing state-of-the-art methods on personalized generation with one single reference image. Specifically, we compare the results of pre-trained IP-Adapter, IP-Adapter-FaceID and IP-Adapter-FaceID-Plus with our InstantID. From Figure 5, there are two substantial observations: 1) ID embedding is fundamental to the task of face preservation. Unlike the CLIP embedding used in IPA, which offers only a coarse representation of images, ID embedding inherently carries rich semantic information, such as identity, age, and gender. This high-level semantic content is crucial for accurate and detailed face preservation. 2) Introducing ID embedding solely at the cross-attention level, as done in IPA-FaceID and IPA-FaceID-Plus, certainly enhances face fidelity. However, this approach can inadvertently impact text control. A notable example is the inability to seamlessly integrate the facial area with the background style. This limitation highlights a trade-off between face fidelity and text control. In contrast, our InstantID method demonstrates robust handling of different reference images, achieving higher fidelity in results. More importantly, it effectively maintains the integrity of the ID, blending it seamlessly into various styles. This comparison underscores InstantID's superiority in preserving identity while maintaining stylistic flexibility and control.

**With LoRAs.** To further demonstrate the superiority of InstantID, we conducted a comparative analysis with LoRA models that were trained using multiple reference images. For this purpose, we selected several pre-trained character LoRA models, such as those resembling Jackie Chan[2] and Emma Watson[3], from

---

[2] https://civitai.com/models/78839/a-man-who-resembles-jackie-chan
[3] https://civitai.com/models/182550/essenz-emma-watson-celebrity-lora-for-sdxl-10

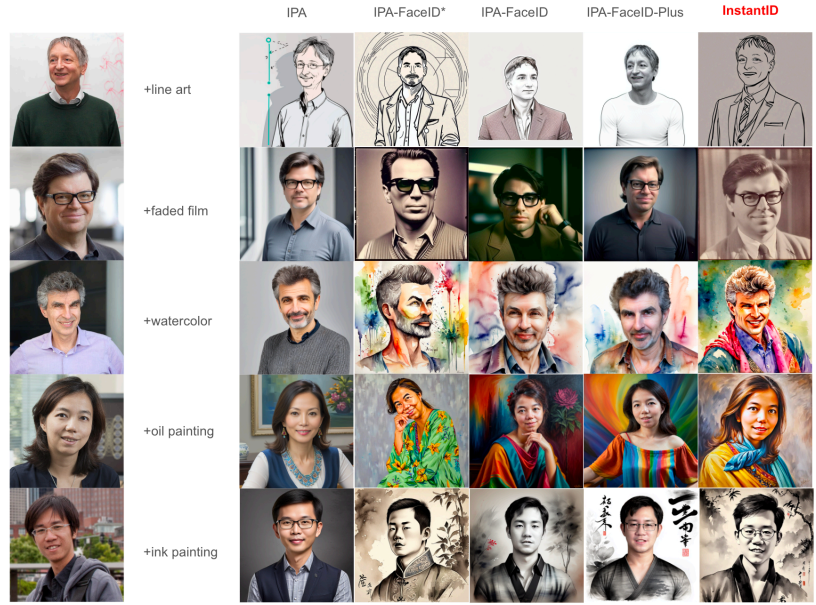

**Fig. 5. Comparison of InstantID with other methods conditioned on different characters and styles.** From left to right are IP-Adapter-SDXL, IP-Adapter-SDXL-FaceID (* indicates experimental version), IP-Adapter-SD1.5-FaceID, IP-Adapter-SD1.5-FaceID-Plus. As shown in the figure, we found that the IP-Adapter that relies on CLIP embedding cannot achieve facial fidelity, and also leads to the degradation of prompt control to generate styles. IP-Adapter-FaceID introduces face embedding, which improves face fidelity, but it still cannot achieve high fidelity. IP-Adapter-FaceID-Plus combines the embedding of face and CLIP, which can achieve good facial fidelity, but there is a style degradation problem, resulting in the face being unable to blend into the background style. In contrast, our proposed InstantID is compatible with various styles while maintaining high fidelity.

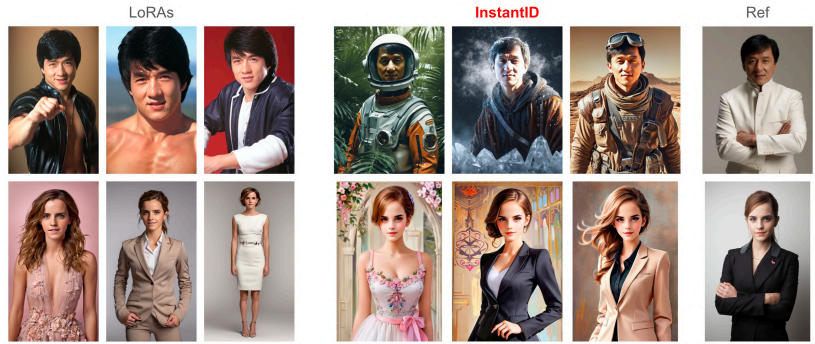

**Fig. 6. Comparison of InstantID with pre-trained character LoRAs.** We can achieve competitive results as LoRAs without any training.

Civitai. The comparison yielded a pleasantly surprising result: InstantID is capable of achieving competitive results using just a single image, even without the necessity of test-time fine-tuning. This finding is significant when contrasted with the LoRA models. LoRA models, on the one hand, have specific requirements regarding data quality and quantity, which can limit their applicability in real-world scenarios. On the other hand, they necessitate the additional training of customized models. In stark contrast, InstantID enables efficient single-step inference with a solitary image, showcasing remarkable efficiency that can greatly benefit various downstream applications. This efficiency, coupled with the lack of need for extensive training or multiple reference images, positions InstantID as a highly practical and versatile tool in the realm of image generation.

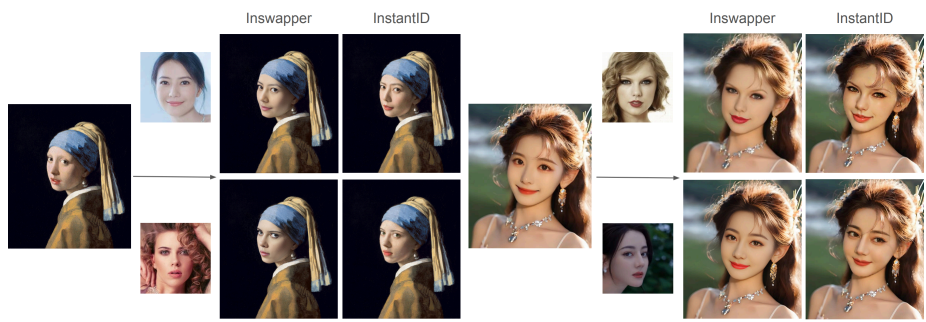

**Fig. 7. Comparison of InstantID with InsightFace Swapper.** However, in non-realistic style, our work is more flexible on the integration of face and background.

In addition, we compare InstantID with InsightFace Swapper[4], a widely used black-box face-swapping model with a non-diffusion model structure, known for its integration into community projects like ROOP[5] and ReActor[6]. Our comparison, as depicted in Figure 7, reveals that while InsightFace Swapper is effective in most standard scenarios, InstantID demonstrates greater flexibility in blending faces with backgrounds, especially in non-realistic styles like anime. This adaptability showcases the advanced capabilities of InstantID in handling a diverse range of artistic expressions.

### 4.4  Applications

Furthermore, InstantID's low-cost, plug-and-play character ID preservation opens the door to a multitude of downstream applications. To spark ideas and showcase the potential, we explore several intriguing use cases. These include novel view synthesis, where InstantID enables the creation of new perspectives and angles while maintaining character consistency. We also delve into identity interpolation between different characters, demonstrating InstantID's ability to seamlessly blend features from multiple identities. Additionally, we highlight the multi-identity synthesis, illustrating how InstantID can manage complex scenarios involving multiple characters. Each of these applications underscores InstantID's versatility and effectiveness in various creative and practical contexts. Results are in Appendix (see Figure 10, Figure 11 and Figure 12, respectively).

---

[4] https://github.com/haofanwang/inswapper
[5] https://github.com/s0md3v/roop
[6] https://github.com/Gourieff/sd-webui-reactor

## 5  Conclusions and Future Work

In this paper, we propose InstantID as a solution for zero-shot identity-preserving generation with a simple plug-and-play module, enabling it to adeptly handle image personalization in any style using only one facial image while maintaining high fidelity. There are two core designs in our InstantID. An Image Adapter that enhances facial detail fidelity and an IdentityNet that ensures strong ID control to preserve complex facial features. Our experimental results affirm the robustness and compatibility of InstantID, alongside its ability to preserve the text editing capabilities of the original model. Its pluggable nature further allows seamless integration with other models such as LoRAs and ControlNet, broadening its applicability without incurring additional costs. The exceptional performance and efficiency of InstantID pave the way for its application in diverse real-world scenarios where identity preservation is paramount. However, despite the effectiveness of InstantID, certain challenges remain to be addressed. The ID embedding in our model, while rich in semantic information like gender and age, has highly coupled facial attributes, which poses a challenge for face editing. Future developments might involve decoupling these facial attribute features to enhance flexibility. Moreover, we may observe some limitations in InstantID associated with the biases inherent in the face model we used. Furthermore, ethical considerations surrounding the maintenance of human faces in our model, particularly the potential creation of offensive or culturally inappropriate imagery, are crucial concerns that warrant careful consideration in future work.

# A  Supplementary Details

## A.1  Implementation Detail

In Figure 3, the spatial control, canny image (b) and depth image (c) are extracted from a same reference pose image Figure 8 (a).

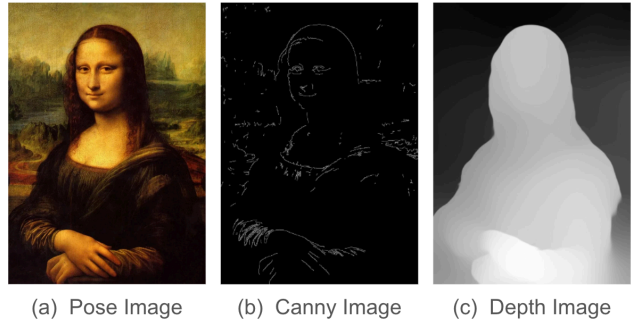

(a) Pose Image    (b) Canny Image    (c) Depth Image

**Fig. 8.** The spatial controls extracted from a reference image.

## A.2  Supplementary Results

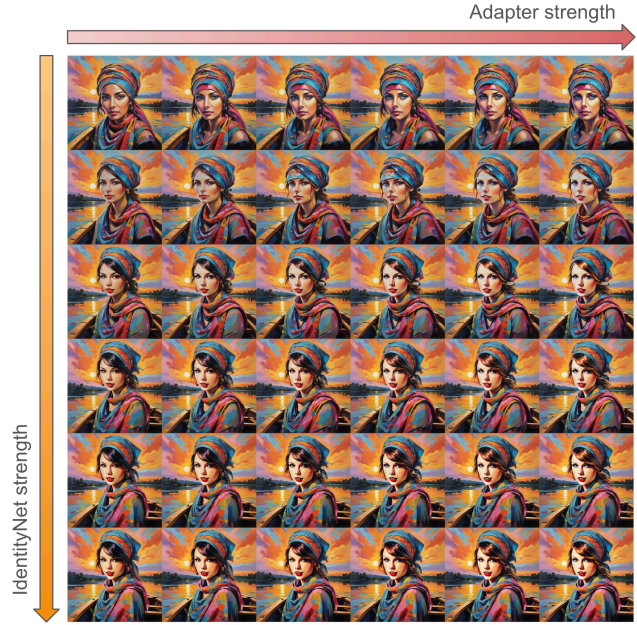

**Fig. 9. Effect of Image Adapter and IdentityNet.** The x-axis and y-axis correspond to the weight strength of Image Adapter and IdentityNet respectively.

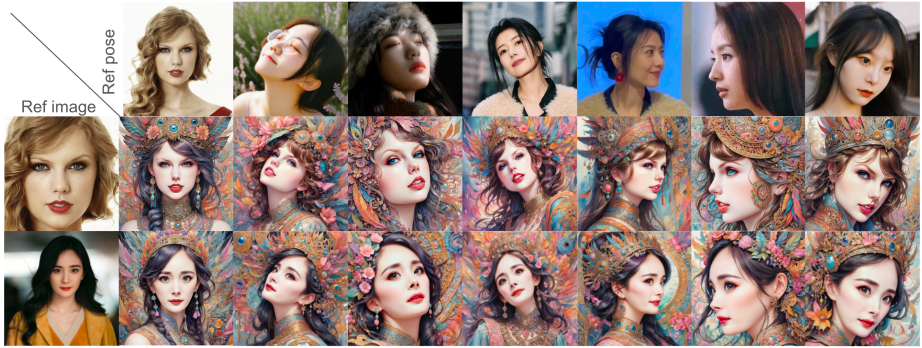

**Fig. 10.** Novel View Synthesis under any given pose.

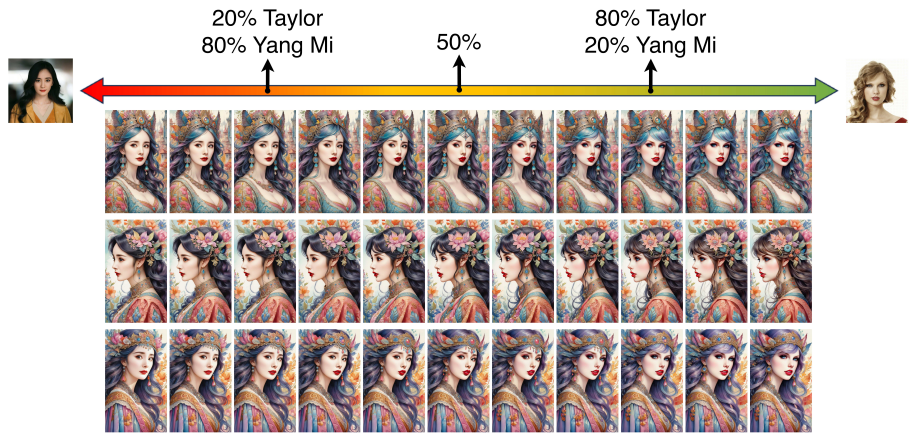

**Fig. 11.** Identity interpolation between different characters.

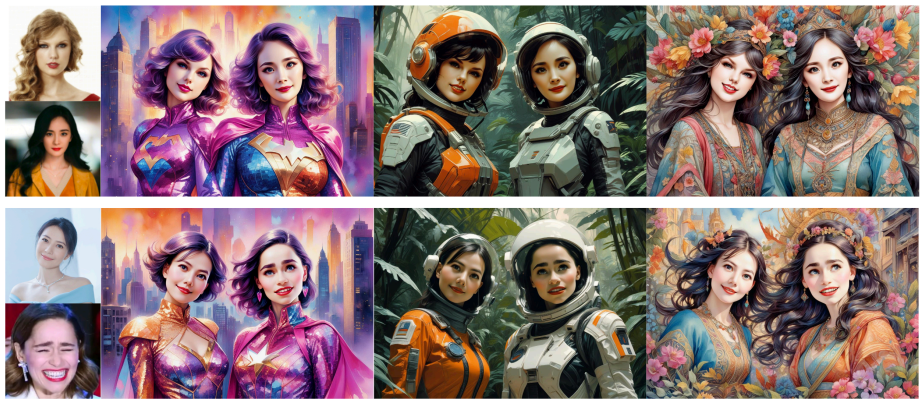

**Fig. 12.** Multi-identity synthesis with regional control.